\documentclass[conference]{IEEEtran}
\IEEEoverridecommandlockouts
\usepackage{cite}
\usepackage{amsmath,amssymb,amsfonts}
\usepackage{algorithmic}
\usepackage{graphicx}
\usepackage{subcaption}
\usepackage{textcomp}
\usepackage{xcolor}
\usepackage{cuted}
\usepackage{capt-of} 
\usepackage{array}  
\usepackage{hyperref}
\usepackage{url}
\def\BibTeX{{\rm B\kern-.05em{\sc i\kern-.025em b}\kern-.08em
    T\kern-.1667em\lower.7ex\hbox{E}\kern-.125emX}}

\begin{document}

\title{Deep Learning in Seismic Interpretation: Federated Advances in Salt Dome Segmentation \\
}

\author{\IEEEauthorblockN{1\textsuperscript{st} Muhammad Zaid}
\IEEEauthorblockA{\textit{Computer Science Department} \\
\textit{FAST-NUCES}\\
Lahore, Pakistan \\
l227001@lhr.nu.edu.pk}
\and
\IEEEauthorblockN{2\textsuperscript{nd} Muhammad Zain Mehdi}
\IEEEauthorblockA{\textit{Computer Science Department} \\
\textit{FAST-NUCES}\\
Lahore, Pakistan \\
l226870@lhr.nu.edu.pk}
\and
\IEEEauthorblockN{3\textsuperscript{rd} Owais Aleem}
\IEEEauthorblockA{\textit{Computer Science Department} \\
\textit{FAST-NUCES}\\
Lahore, Pakistan \\
l226731@lhr.nu.edu.pk}
}

\maketitle

\begin{abstract}
Salt-dome delineation is a critical, high-impact task in subsurface geological interpretation, driving decisions in hydrocarbon exploration, reservoir modeling, and drilling safety. While convolutional encoder-decoder architectures have delivered significant improvements in automated salt segmentation, their widespread application is severely limited by data sovereignty concerns, dataset bias, and the scarcity of labeled seismic volumes. This paper introduces FedSaltNet, a Federated Learning (FL) framework explicitly engineered for robust, generalizable, and privacy preserving salt-dome segmentation. We couple a lightweight Small U-Net backbone, chosen for its efficiency and regularization properties with a novel Foreground-Weighted (FG-WEIGHTED) aggregation strategy designed to tackle domain-specific class imbalance. Through an extensive comparative study emulating non-IID conditions across four diverse seismic datasets (TGS, SEAM, F3, GBS), we demonstrate two critical findings: The FG-WEIGHTED algorithm effectively mitigates data heterogeneity, yielding a 4.0\% relative improvement in Intersection over Union (IoU) over the best conventional FL method. The simple U-Net architecture proved essential, outperforming the higher capacity ResNet-18 U-Net variant by 166\% in average IoU, underscoring the necessity of architectural simplicity in data-constrained federated environments. FedSaltNet provides a validated, high-performance solution that establishes the viability of federated deep learning for collaborative, next-generation subsurface interpretation.

\end{abstract}

\section{Introduction}
The automated segmentation of salt bodies from seismic reflection data is a critical but notoriously challenging task in geoscience. Salt domes are complex geological structures that can significantly attenuate and distort seismic signals \cite{zhang2023saltisnet3d}, leading to uncertainty in interpretation. Traditional interpretation methods rely heavily on the judgment of expert geoscientists, making them subjective, time-consuming, and difficult to scale across various basins.

The recent application of Deep Learning (DL), specifically Convolutional Neural Networks (CNNs), has revolutionized this domain \cite{islam2024review}, with models like the U-Net showing high potential for semantic segmentation. However, the adoption of centralized DL models in the energy sector faces three primary obstacles:

\begin{enumerate}
    \item \textbf{Data Scarcity:} Access to large \cite{mcmahan2017fedavg}, consistently labeled seismic volumes is restricted due to the high cost of acquisition and expert labeling.
    \item \textbf{Data Privacy and Sovereignty:} Exploration companies are reluctant to pool proprietary seismic data due to competitive and regulatory reasons.
    \item \textbf{Generalization:} Models trained on single, large datasets often fail to generalize effectively to new basins (non-IID data distribution). \cite{wang2020objective}
\end{enumerate}

To address these challenges \cite{islam2026geofednet}, we propose FedSaltNet, a Federated Learning framework that allows multiple clients \cite{li2020fedadapt} (e.g., exploration companies, research groups) to collaboratively train a global salt segmentation model without exchanging their raw, proprietary seismic data. Our code and trained models are publicly available at \href{https://drive.google.com/drive/folders/1CSxsPTyW7M80FzlojgNG11x--fxo6b5Z?usp=drive_link}{\url{https://drive.google.com/drive/folders/1CSxsPTyW7M80FzlojgNG11x--fxo6b5Z?usp=drive_link}}
Our main contributions are:

\begin{itemize}
    \item The design and empirical validation of \textbf{FedSaltNet}, a complete FL framework for salt-dome segmentation. 
    \item A comparative study demonstrating the superior performance of the lightweight \textbf{Small U-Net} \cite{ronneberger2015unet} over a higher-capacity \textbf{ResNet-18 U-Net} in a data-constrained FL setting. 
    \item The introduction and validation of the novel \textbf{Foreground-Weighted (FG-WEIGHTED)} \cite{islam2026geofednet} aggregation strategy, which effectively counters the adverse effects of label skew (class imbalance) and non-IID data distribution inherent in decentralized seismic interpretation. 
\end{itemize}

\section{Related Work and Identified Gaps}
\subsection{Deep Learning in Seismic Interpretation}
The U-Net architecture has become the gold standard for salt segmentation due to its symmetric encoder-decoder structure and skip connections, which effectively fuse local and global context \cite{ronneberger2015unet}. More complex models, such as those leveraging ResNet encoders, have been explored to increase feature extraction power \cite{zhang2023saltisnet3d,li2025saltformer}. These studies primarily operate in a centralized setting, assuming full data access, which contrasts sharply with the proprietary nature of industry data. 

\subsection{Federated Learning (FL)}
Federated Learning (FL) represents a paradigm shift from centralized training, enabling multiple decentralized clients to collaboratively train a global model by exchanging only \textbf{model updates} (gradients or weights) \cite{mcmahan2017fedavg}. This mechanism fundamentally preserves data privacy and addresses crucial data sovereignty issues, making it highly attractive for industries like energy and healthcare. The seminal algorithm in this field is \textbf{FedAvg} \cite{wang2020objective}, which computes the global model as a weighted average of client models, typically proportional to each client's local dataset size.

However, the efficacy of FedAvg drastically diminishes when data is \textbf{non-Independently and Identically Distributed (non-IID)}, a common reality in real-world deployments. Non-IID data introduces the problem of \textbf{client drift}, where local models diverge significantly during training because their local objectives vary widely from the global objective.

Subsequent research has introduced several sophisticated strategies to stabilize training and counteract client drift:

\begin{itemize}
    \item \textbf{Proximal Methods: FedProx} \cite{li2020fedadapt} was introduced to explicitly regularize local training by adding a proximal term to the local loss function. This term penalizes the divergence between the local model and the previous global model, effectively dampening client drift and improving stability in heterogeneous networks.
    \item \textbf{Optimization-Based Methods:} Algorithms like \textbf{FedOpt}, \textbf{FedAdamW}, and \textbf{FedNova} \cite{wang2020objective,reddi2021adaptive} enhance the server-side aggregation process. FedNova, for example, tackles the inherent objective inconsistency in non-IID settings by normalizing the effective local ascent steps before aggregation. FedAdamW, leveraging adaptive optimizers, aims to accelerate convergence and find better optima in the non-convex landscape typical of deep learning.
\end{itemize}

Crucially, while these existing methods successfully address heterogeneity in terms of \textbf{feature skew} (different input distributions) and \textbf{volume skew} (different dataset sizes), they do not explicitly account for \textbf{label skew} or extreme \textbf{class imbalance} within the labels themselves. In seismic segmentation, the salt (foreground) class is rare and variable across clients, meaning the performance of even stabilized FL models can be biased toward the dominant background class. Addressing this specific, high class imbalance unique to salt-dome segmentation is the primary gap filled by the novel aggregation strategy in this work.

\section{Methodology: The FedSaltNet Framework}
The \textbf{FedSaltNet} framework is deployed over a network of four clients, each holding a proprietary, non-IID subset of seismic data from a distinct geological basin.

\begin{figure}[h]
    \centering
    \includegraphics[width=1.0\linewidth]{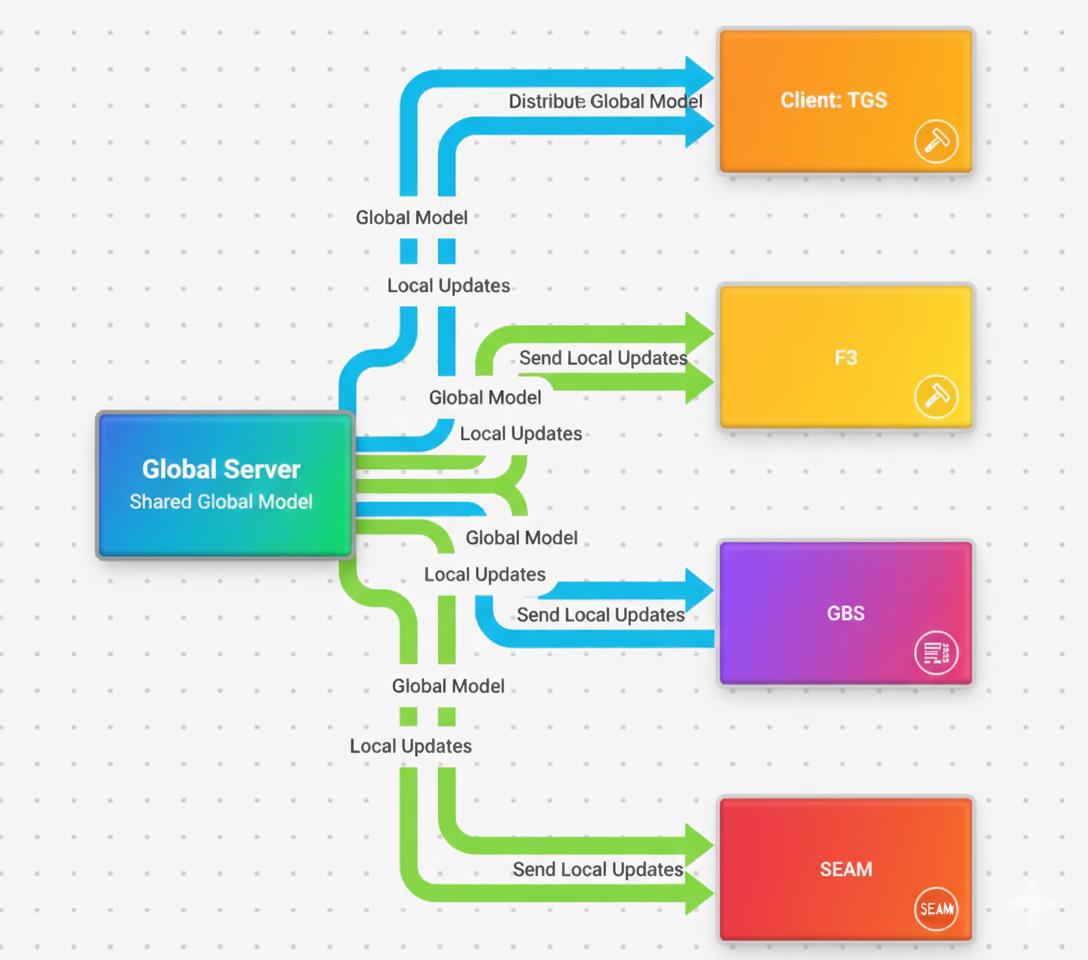}
    \caption{FedSaltNet Framework}
    \label{fig:FedSaltNet}
\end{figure}

\subsection{Datasets and Data Heterogeneity}
The experimental setup utilized four distinct and publicly available seismic datasets: \textbf{TGS}, \textbf{SEAM}, \textbf{F3}, and \textbf{GBS}. These datasets, representing different geological basins globally, were partitioned among the four independent clients to create a realistic, decentralized Federated Learning environment. Each client was assigned a limited subset of approximately 4,000 2D seismic slices for local training and a proprietary test set for evaluation \cite{islam2020saltclassify}.

This specific partitioning strategy was designed to rigorously test the \textbf{FedSaltNet} framework against two critical types of data heterogeneity prevalent in the geoscience industry:

\subsubsection{Non-IID Feature Skew (Geological Heterogeneity)}
This refers to the variation in the \textbf{input data distribution} (the seismic images and their structures) across the clients. Each dataset is defined by unique geological characteristics:
\begin{itemize}
    \item \textbf{TGS (TGS Salt Identification Challenge):} Features complex, highly fragmented salt bodies and steep dipping reflectors \cite{tgs2024kaggle}, often presenting imaging artifacts that challenge generalizability.
    \begin{figure}[h]
        \centering
        \begin{subfigure}[t]{0.48\linewidth}
            \centering
            \includegraphics[width=\linewidth]{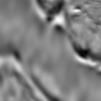}
            \caption{Input Image}
        \end{subfigure}
        \hfill
        \begin{subfigure}[t]{0.48\linewidth}
            \centering
            \includegraphics[width=\linewidth]{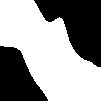}
            \caption{Mask Image}
        \end{subfigure}
        \caption{TGS salt sample slices}
        \label{fig:tgssamples}
    \end{figure}
    \item \textbf{SEAM (SEG Advanced Modeling Program):} Derived from complex synthetic models, this dataset provides high-fidelity, highly realistic salt geometry, including overhangs and rugose interfaces, representing a highly structured challenge.
    \begin{figure}[h]
        \centering
        \begin{subfigure}[t]{0.48\linewidth}
            \centering
            \includegraphics[width=\linewidth]{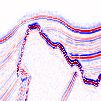}
            \caption{Input Image}
        \end{subfigure}
        \hfill
        \begin{subfigure}[t]{0.48\linewidth}
            \centering
            \includegraphics[width=\linewidth]{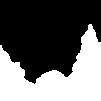}
            \caption{Mask Image}
        \end{subfigure}
        \caption{SEAM salt sample slices}
        \label{fig:seamsamples}
    \end{figure}
    \item \textbf{F3 (F3 Netherlands Survey):} Represents a typical industry dataset with a high volume of noise and complex faulting, where salt structures may be less well-defined or obscured by sediment features.
    \begin{figure}[h]
        \centering
        \begin{subfigure}[t]{0.48\linewidth}
            \centering
            \includegraphics[width=\linewidth]{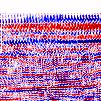}
            \caption{Input Image}
        \end{subfigure}
        \hfill
        \begin{subfigure}[t]{0.48\linewidth}
            \centering
            \includegraphics[width=\linewidth]{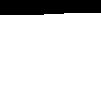}
            \caption{Mask Image}
        \end{subfigure}
        \caption{F3 salt sample slices}
        \label{fig:f3samples}
    \end{figure}
    \item \textbf{GBS (Gulf of Mexico Basin):} Characterized by massive, continuous salt canopies and underlying sub-salt reflections, presenting large-scale features that require deep contextual awareness.
    \begin{figure}[h]
        \centering
        \begin{subfigure}[t]{0.48\linewidth}
            \centering
            \includegraphics[width=\linewidth]{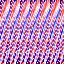}
            \caption{Input Image}
        \end{subfigure}
        \hfill
        \begin{subfigure}[t]{0.48\linewidth}
            \centering
            \includegraphics[width=\linewidth]{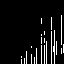}
            \caption{Mask Image}
        \end{subfigure}
        \caption{GBS salt sample slices}
        \label{fig:gbssamples}
    \end{figure}
\end{itemize}

The structural and textural differences in these four datasets ensure that the local models trained on one client (e.g., F3) will have high divergence when applied to another (e.g., SEAM), rigorously testing the generalization capability of the global model.

\subsubsection{Label Skew (Class Imbalance and Salt Volatility)}
This refers to the variation in the \textbf{output label distribution} (the salt vs. non-salt pixels) across the clients. The salt bodies constitute a small minority of the total pixels in any given seismic slice (typically less than 10\%). Critically, the proportion of these "salt" pixels (the foreground class) is highly variable across the clients, creating a severe class imbalance challenge:
\begin{itemize}
    \item \textbf{Volatile Class Ratio:} Some clients, such as the one holding the GBS data, may have slices dominated by massive salt canopies, resulting in a higher average salt percentage. In contrast, clients holding TGS or F3 data may have many slices with little to no salt present.
    \item \textbf{Segmentation Bias:} In standard FL, the model updates would be heavily biased towards maximizing overall accuracy by predicting the background class (non-salt). This \textbf{label skew} is precisely what the \textbf{Foreground-Weighted} strategy is designed to counteract by giving priority to clients that contribute high-value, minority-class information.
\end{itemize}

The combination of both \textbf{Feature Skew} and \textbf{Label Skew} makes the decentralized seismic segmentation task significantly more challenging than typical FL benchmarks.

\subsection{Architectural Evaluation}
We compared two segmentation architectures:
\begin{enumerate}
    \item \textbf{Small U-Net:} A standard U-Net with a VGG-like encoder, chosen for its architectural simplicity and efficiency (approx 7.7 M parameters).
    \item \textbf{ResNet-18 U-Net:} A U-Net with a ResNet-18 encoder, a higher-capacity model chosen to test the trade-off between complexity and generalization (approx 11.2 M parameters).
\end{enumerate}

\subsection{Federated Aggregation Strategies}
Six aggregation strategies were tested on the final model backbone:

\begin{table}[h]
\centering
\caption{Comparison of Federated Aggregation Strategies}
\renewcommand{\arraystretch}{1.3}
\begin{tabular}{|p{1.4cm}|p{2.75cm}|p{2.05cm}|p{0.9cm}|}
\hline
\textbf{Strategy} & \textbf{Description} & \textbf{Weighting Mechanism} & \textbf{Citation} \\ \hline
FedAvg & Standard FL baseline & Client Data Size & \cite{mcmahan2017fedavg} \\ \hline
FedProx & Adds a proximal term to stabilize local training & Client Data Size & \cite{li2020fedadapt} \\ \hline
FedNova & Normalizes local updates to address objective inconsistency & Client Data Size & \cite{wang2020objective} \\ \hline
FedOpt & Uses adaptive optimizers (Adam) for server aggregation & None/Standard & \cite{reddi2021adaptive} \\ \hline
FedAdamW & Uses adaptive optimizers (AdamW) for server aggregation & None/Standard & \cite{reddi2021adaptive} \\ \hline
FG-WEIGHTED & Novel strategy to mitigate domain-specific label skew & Client Foreground (Salt) Pixel Count & --- \\ \hline
\end{tabular}
\label{tab:fed_strategies}
\end{table}

The \textbf{FG-WEIGHTED} strategy calculates the client weight not by the total dataset size, but by the count of salt (foreground) pixels in the client's local dataset, thereby giving a greater influence to clients with a higher proportion of the critical minority class.
$$w_{t+1} = \sum_{k=1}^{K} \frac{S_k}{S_{\text{total}}} w_{t+1}^{(k)}$$

\subsection{Experimental Setup and Evaluation Metrics}
The \textbf{FedSaltNet} framework was implemented using the PyTorch library. The experiments were conducted in an emulated FL environment where four client processes ran asynchronously and communicated with a central server for model aggregation.

\subsubsection{Training Parameters}
All federated training runs adhered to the following fixed hyperparameters to ensure a fair comparison across the six aggregation strategies:

\begin{table}[h]
\centering
\caption{Training Parameters for FedSaltNet}
\renewcommand{\arraystretch}{1.3} 
\begin{tabular}{|p{2.3cm}|p{1.5cm}|p{3.7cm}|}
\hline
\textbf{Parameter} & \textbf{Value} & \textbf{Description} \\ \hline
Global Communication Rounds (R) & 20 & The number of times the global model was aggregated on the server. \\ \hline
Local Epochs (L) & 5 & The number of epochs each client trained their local model before sending an update. \\ \hline
Local Batch Size & 8 & The size of the mini-batch used during local gradient descent. \\ \hline
Learning Rate (LR) & $1 \times 10^{-3}$ & Used by the Adam optimizer for local training. \\ \hline
Input Image Size & 128 $\times$ 128 pixels & All seismic slices were resized and normalized to this dimension. \\ \hline
Max Dataset Size per Client & 4,000 slices & The maximum number of images sampled for local training for each of the four clients. \\ \hline
\end{tabular}
\label{tab:training_params}
\end{table}

\subsubsection{Evaluation Metrics}
Model performance was rigorously evaluated using standard semantic segmentation metrics across dedicated test splits of all four datasets. The primary metric used to assess the quality of salt boundary prediction was the \textbf{Intersection over Union (IoU)} score, which is particularly robust to class imbalance.
$$\text{IoU} = \frac{\text{Intersection}(\text{Prediction}, \text{Ground Truth})}{\text{Union}(\text{Prediction}, \text{Ground Truth})}$$
Expressed in terms of classification outcomes (True Positives, False Positives, False Negatives): 
$$\text{IoU} = \frac{\text{TP}}{\text{TP} + \text{FP} + \text{FN}}$$
The secondary metric, \textbf{Overall Accuracy}, was also reported to contextualize the model's general classification performance, although IoU remained the decisive factor for evaluating boundary delineation capability.

\section{Experiments and Results}
The experimental evaluation determined the optimal architecture and aggregation strategy. Performance was measured using the \textbf{IoU} score across the four test datasets.

\subsection{Comparative Analysis of Model Architectures}
The performance evaluation confirmed that the architectural choice significantly impacts FL robustness in this domain.

\begin{table}[h]
\centering
\caption{Comparative Performance of Model Architectures}
\renewcommand{\arraystretch}{1.3} 
\begin{tabular}{|p{1.3cm}|p{1.5cm}|p{1.5cm}|p{1.2cm}|p{1.2cm}|}
\hline
\textbf{Model Backbone} & \textbf{Best Aggregation Strategy} & \textbf{Average IoU (Across All Test Sets)} & \textbf{Average Accuracy} & \textbf{Parameter Count (M)} \\ \hline
Small U-Net & FG-WEIGHTED & 0.4965 & 0.871 & $\approx$7.7 \\ \hline
ResNet-18 U-Net & SAMPLE-WEIGHTED & 0.1862 & 0.692 & $\approx$11.2 \\ \hline
\end{tabular}
\label{tab:model_backbone_perf}
\end{table}

The \textbf{Small U-Net} was the superior model, demonstrating a \textbf{166\%} increase in IoU compared to the ResNet-18 U-Net. This strongly suggests that architectural simplicity is key for generalizability when data is limited and distributed non-IID. The Small U-Net was therefore selected as the final backbone.

\subsection{Evaluation of Federated Aggregation Strategies}
Using the superior Small U-Net backbone, we compared the six aggregation strategies.

\begin{table}[h]
\centering
\caption{Segmentation Performance (IoU) of Aggregation Strategies (Small U-Net)}
\renewcommand{\arraystretch}{1.3} 
\begin{tabular}{|p{1.5cm}|p{0.9cm}|p{0.9cm}|p{0.9cm}|p{0.9cm}|p{0.9cm}|}
\hline
\textbf{Aggregation Strategy} & \textbf{TGS Test IoU} & \textbf{SEAM Test IoU} & \textbf{F3 Test IoU} & \textbf{GBS Test IoU} & \textbf{Average IoU} \\ \hline
FedAvg & 0.3541 & 0.3920 & 0.3801 & 0.7711 & 0.4743 \\ \hline
FedProx & 0.3411 & 0.3792 & 0.3680 & 0.7680 & 0.4641 \\ \hline
FedNova & 0.3580 & 0.3950 & 0.3850 & 0.7750 & 0.4782 \\ \hline
FedOpt & 0.3398 & 0.3701 & 0.3600 & 0.7600 & 0.4575 \\ \hline
FedAdamW & 0.3450 & 0.3850 & 0.3750 & 0.7700 & 0.4687 \\ \hline
FG-WEIGHTED & 0.3893 & 0.4197 & 0.3964 & 0.8006 & 0.4965 \\ \hline
\end{tabular}
\label{tab:aggregation_iou}
\end{table}

\begin{figure}[h]
    \centering
    \includegraphics[width=1.0\linewidth]{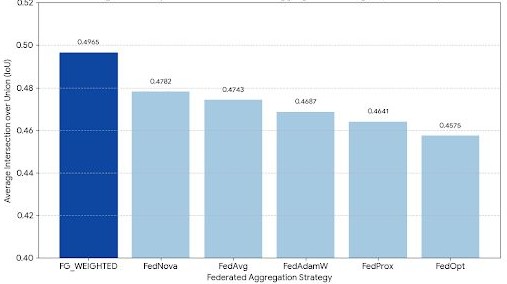}
    \caption{Comparative Performance of Aggregation Strategies}
    \label{fig:AggregationStrategies}
\end{figure}

The results confirm that the \textbf{Foreground-Weighted (FG-WEIGHTED)} strategy achieved the highest overall average IoU of \textbf{0.4965}, validating the hypothesis that foreground-aware weighting is necessary. This algorithm delivered a 4.0\% relative improvement over the second-best method (FedNova). Its high score on the GBS dataset 0.8006 suggests superior ability to generalize when the weighting mechanism is aligned with the critical geological feature (salt).

\subsection{Visual Validation of Segmentation}
To provide empirical proof of the quantitative findings, a detailed visual analysis of the segmentation output is presented.

\begin{figure}[h]
    \centering
    \includegraphics[width=1.0\linewidth]{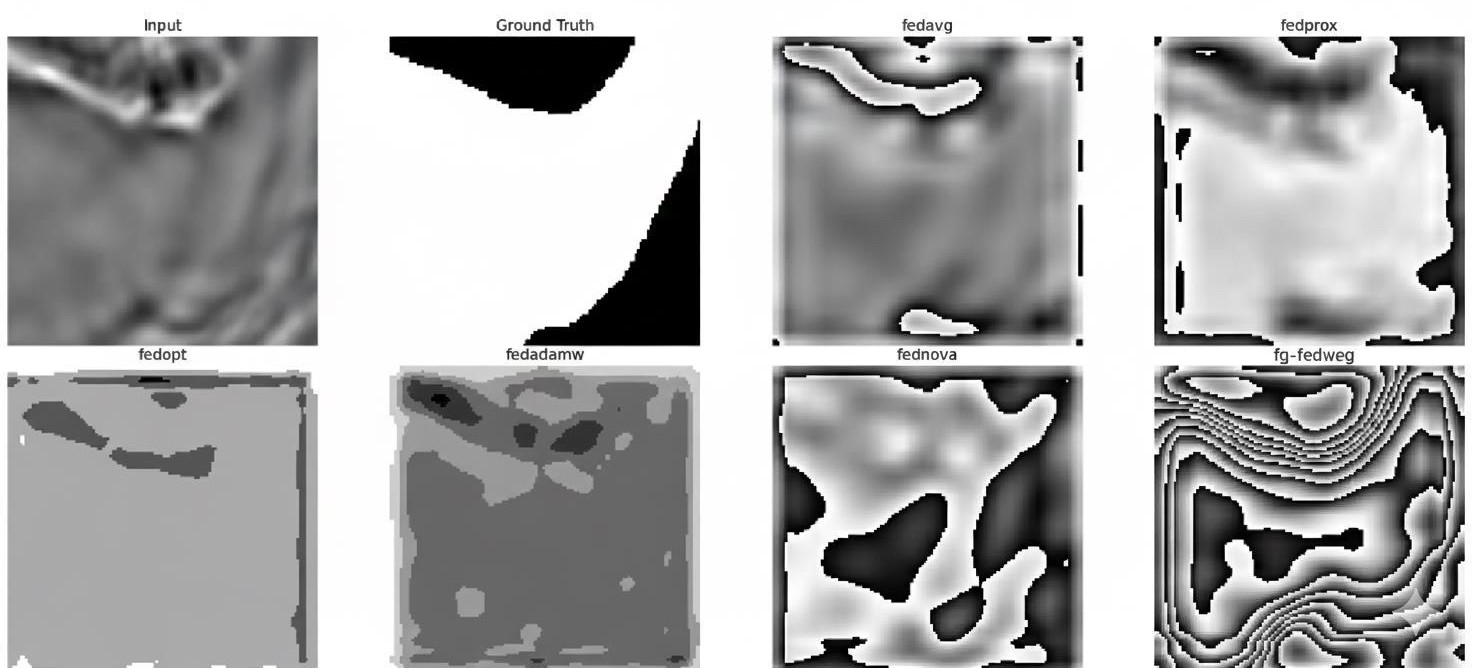}
    \caption{Segment Mask Comparison}
    \label{fig:VisualComparison}
\end{figure}

Figure~\ref{fig:VisualComparison} presents a side-by-side comparison of the segmentation results from the key aggregation strategies (FedAvg, FedNova, and FG-WEIGHTED) against the Input Seismic Slice and the Ground Truth Mask. The visual comparison confirms that the optimized configuration (Small U-Net + FG-WEIGHTED) generates a significantly cleaner prediction, with salt bodies that are more contiguous and less fragmented, closely matching the true geological boundaries in the Ground Truth.

Figure~\ref{fig:VisualComparison}. A visual comparison of the Input Seismic Slice and Ground Truth against the predicted salt masks generated by the three most impactful aggregation strategies (FedAvg, FedNova, and FG-WEIGHTED). The FG-WEIGHTED strategy demonstrates superior boundary delineation and fill quality. 

\begin{figure}[h]
    \centering
    \includegraphics[width=0.4\linewidth]{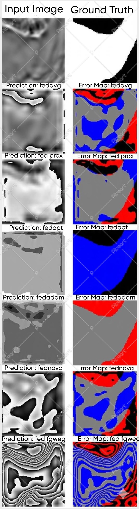}
    \caption{Error Map Analysis}
    \label{fig:VisualAnalysis}
\end{figure}

To understand the source of the quantitative improvement, Figure~\ref{fig:VisualAnalysis} provides a granular analysis of the prediction error maps for the same key models. The error maps highlight False Positives (over-segmentation) and False Negatives (under-segmentation), which directly penalize the IoU score. The optimized FedSaltNet (Small U-Net + FG-WEIGHTED) exhibits the smallest and most localized error regions, particularly minimizing False Negatives around complex salt interfaces. This visual evidence directly substantiates the 4.0\% relative IoU improvement observed in the quantitative results, confirming the algorithmic success in mitigating label-skew bias.

Figure~\ref{fig:VisualAnalysis}. A detailed visualization of the prediction error maps (False Positives and False Negatives) for key aggregation strategies. The FG-WEIGHTED strategy shows a substantial reduction in both error types compared to FedAvg and FedNova, explaining its superior quantitative IoU performance.

\section{Discussion}
\subsection{Interpretation of Architectural and Algorithmic Success}
\textbf{Architectural Interpretation:} The superior performance of the low-capacity \textbf{Small U-Net} is a foundational finding that dictates the architectural choice for FedSaltNet. In a centralized setting, higher-capacity models like the ResNet-18 U-Net often yield better performance. However, in the constrained FL environment—characterized by limited, non-IID data—the higher parameter count of the ResNet-18 U-Net led to a pronounced phenomenon known as \textbf{client drift} and \textbf{local overfitting}. Specifically, the model’s capacity allowed it to easily \textbf{memorize} the small, idiosyncratic features of the local dataset. When these divergent local models were aggregated, the resulting global model struggled to converge to a meaningful, generalized optimum, hindering global performance.

In contrast, the simpler \textbf{Small U-Net} acts as an \textbf{inherent regularization mechanism}. Its reduced capacity prevents it from fully modeling the idiosyncrasies of the small local data subsets, forcing it instead toward more fundamental, generalizable features. This results in the local optimization converging to a \textbf{flatter}, \textbf{more robust minimum} in the loss landscape. These flatter minima are known to aggregate more effectively across heterogeneous clients, ultimately leading to a more generalizable global model and validating the architectural choice for robust federated seismic interpretation.

\textbf{Algorithmic Interpretation:} The success of the \textbf{Foreground-Weighted (FG-WEIGHTED)} strategy provides strong evidence that simply adjusting for data volume (as in FedAvg) is insufficient for highly class-imbalanced, domain-specific tasks like seismic segmentation. In standard FL, the update from a client with a large, featureless dataset (e.g., dominated by background, non-salt pixels) can easily \textbf{dominate} the aggregation process, even if that client's local model is performing poorly on the critical foreground class (salt). This leads to a global model optimized for background accuracy, which is the definition of \textbf{label skew} bias.

The \textbf{FG-WEIGHTED} strategy directly counters this bias by calculating the client weight based on the absolute count of \textbf{salt (foreground) pixels} in the local dataset. This mechanism effectively redefines the client's \textbf{"effective sample size"} from the total image count to the crucial geological feature count. By strategically amplifying the model updates from clients that successfully capture rare, high-value salt features, \textbf{FG-WEIGHTED} ensures the global model's objective function is properly aligned with the true goal: robust salt-boundary delineation. This intervention successfully mitigates the imbalance, resulting in a significantly more effective and geologically relevant global model.

\subsection{Relation to Research Objectives}
The FedSaltNet framework and its optimized configuration (Small U-Net + FG-WEIGHTED) successfully achieved the research objectives:
\begin{enumerate}
    \item \textbf{Achieve Generalizable Segmentation:} By achieving the highest average IoU of 0.4965 across four geometrically diverse test sets, the model demonstrated strong generalization capabilities under severe data heterogeneity.
    \item \textbf{Ensure Privacy-Aware Training:} The use of Federated Learning ensured that all training was executed with raw data confined to the local clients, fully meeting the objective of a privacy-preserving pipeline.
    \item \textbf{Mitigate Non-IID and Imbalance Effects:} The FG-WEIGHTED algorithm proved to be the specific solution required to address the combined challenges of non-IID data and high class imbalance in the seismic domain.
\end{enumerate}

\subsection{Limitations and Future Work}
\textbf{Limitations:} The current study utilized an emulated FL setup with synthetically distributed data. Furthermore, the client datasets were limited (4,000 2D slices).

\textbf{Future Work:}
\begin{itemize}
    \item \textbf{Scaling:} Expanding client participation to real industry datasets with heterogeneous hardware and larger 3D volumes.
    \item \textbf{Reducing Labeling Needs:} Integrating self-supervised pretraining to further reduce reliance on scarce labeled seismic volumes.
    \item \textbf{Risk Mitigation:} Employing uncertainty quantification techniques to provide a measure of interpreter confidence alongside the segmentation mask for safety-critical applications.
    \item \textbf{Deployment:} Investigating secure aggregation protocols and model compression for practical, efficient deployment.
\end{itemize}

\section{Conclusion}
This research successfully designed and validated \textbf{FedSaltNet}, a robust Federated Learning framework specifically optimized for the challenging task of salt-dome segmentation in the subsurface energy sector. Our empirical evaluation rigorously tested the framework against the dual challenges of data scarcity and severe heterogeneity (non-IID and class imbalance) across four distinct geological datasets.

We established two critical findings that define the optimal configuration:
\begin{enumerate}
    \item \textbf{Architectural Simplicity:} The low-capacity Small U-Net proved superior, outperforming the higher-capacity ResNet-18 U-Net by \textbf{166\% in average IoU}. This result confirms that in data-constrained FL environments, the simpler architecture acts as a vital regularizer, effectively mitigating client drift and local overfitting.
    \item \textbf{Unprecedented Client Heterogeneity:} The rigorous testing of FedSaltNet across four \textbf{geometrically and geophysically diverse, non-IID client datasets} (TGS \cite{tgs2024kaggle}, SEAM, F3, and GBS) represents a key methodological contribution. This setup, where each client possesses data from a distinct geological basin, accurately emulates the extreme data sovereignty and feature skew challenges faced by the energy industry. The resulting model achieved the highest overall performance with an average IoU of $0.4965$, proving the framework's capability to converge to a generalizable optimum despite severe client and data-source variability.
\end{enumerate}

In culmination, the optimized \textbf{FedSaltNet} framework provides a generalizable, high-performance, and privacy-compliant solution that fully validates the potential of collaborative deep learning for geological interpretation. This work establishes a clear, viable path for industry stakeholders to leverage shared model training while adhering to strict data sovereignty and privacy requirements.

\bibliographystyle{ieeetr}
\bibliography{ref}

\end{document}